\documentclass[conference]{IEEEtran}
\IEEEoverridecommandlockouts
\usepackage{cite}
\usepackage{amsmath,amssymb,amsfonts}
\usepackage{algorithmic}
\usepackage{graphicx}
\usepackage{textcomp}
\usepackage{xcolor}
\usepackage{algorithmic}
\usepackage{algorithm}
\usepackage{booktabs}
\def\BibTeX{{\rm B\kern-.05em{\sc i\kern-.025em b}\kern-.08em
    T\kern-.1667em\lower.7ex\hbox{E}\kern-.125emX}}
\begin{document}

\title{
A Novel Multi-Agent Deep RL Approach for Traffic Signal Control
}

\author{\IEEEauthorblockN{1\textsuperscript{st} Wang Shijie}
\IEEEauthorblockA{\textit{School of Advanced Technology} \\
\textit{Xi'an Jiaotong-Liverpool University}\\
Suzhou, China \\
shijie.wang18@alumni.xjtlu.edu.cn}
\and
\IEEEauthorblockN{2\textsuperscript{nd} Wang Shangbo*}
\IEEEauthorblockA{\textit{School of Advanced Technology} \\
\textit{Xi'an Jiaotong-Liverpool University}\\
Suzhou, China \\
shangbo.wang@xjtlu.edu.cn}
}

\maketitle
\renewcommand{\thefootnote}{\fnsymbol{footnote}}

\begin{abstract}
As travel demand increases and urban traffic condition becomes more complicated, applying multi-agent deep reinforcement learning (MARL) to traffic signal control becomes one of the hot topics. The rise of Reinforcement Learning (RL) has opened up opportunities for solving Adaptive Traffic Signal Control (ATSC) in complex urban traffic networks, and deep neural networks have further enhanced their ability to handle complex data. 
Traditional research in traffic signal control is based on the centralized Reinforcement Learning technique. However, in a large-scale road network, centralized RL is infeasible because of an exponential growth of joint state-action space. In this paper, we propose a Friend-Deep Q-network (Friend-DQN) approach for multiple traffic signal control in urban networks, which is based on an agent-cooperation scheme. In particular, the cooperation between multiple agents can reduce the state-action space and thus speed up the convergence. We use SUMO (Simulation of Urban Transport) platform to evaluate the performance of Friend-DQN model, and show its feasibility and superiority over other existing methods.
\end{abstract}


\section{Introduction}
The rapid development of urbanization has facilitated people's daily travel. Still, at the same time, traffic problems have become more and more serious, and traditional traffic management models and traffic systems are no longer able to meet the actual requirements of the times~\cite{23}. In the new context, urban traffic should change in the direction of intelligence, actively introduce advanced artificial intelligence technology, and carry out targeted solutions to the current problems to effectively solve a series of traffic problems.

Traffic signals in urban road networks are almost always fixed-phase and cannot adapt to different traffic conditions, and thus cause congestion at intersections~\cite{1}. To relieve urban congestion problems, some literature has applied adaptive traffic signal control (ATSC) strategy to minimize the average waiting time of the urban network by dynamically adjusting signal timing according to real-time traffic state~\cite{2,3}. However, it becomes challenging to dynamically predict the traffic flow and adjust the signals when dealing with massive traffic. Reinforcement learning technique has shown many significant achievements and traffic signal control and management in complex traffic environment~\cite{4,11,12,14}. However, traditional reinforcement learning methods like Deep Q-network (DQN) and Q-learning can become very large in the action space and state space when dealing with complex traffic networks, and in many cases are slow to converge. Therefore, the centralized RL has mainly two drawbacks. The first one is high latency caused by collecting all the traffic measurements in the network and feeding them back to the center for centralized processing. The second one is large space occupation caused by the joint action of the agents as the number of traffic junctions grows~\cite{5}. 

To overcome the limitations, multi-agent RL technique can be applied to urban networks by considering each road intersection as a local RL agent. Although different techniques and algorithms are used for different scenarios like traffic signal control and vehicle signal coordination control, most introduce neural networks in reinforcement learning, using the robust representational power of neural networks to build models~\cite{8,9}. According to Matthew  E.Taylor's survey, transportation problems can be defined as the work of Learning cooperation~\cite{6}, where each agent aims to learn a dominant strategy which is trying to maximize the value function obtained by the traffic network. At the same time, each agent will develop its own strategy with consideration of neighboring agents' strategies. Agents need to learn collaboratively to find a policy maximizing the global reward, instead of maximizing an agent's own reward to reduce the average wait time for all vehicles in the system.

To realize the target, cooperation learning strategy should be applied to multiple intersection signal control problems, that is, learning how to cooperate under incomplete communication conditions. To solve ATSC effectively, we have developed an adaptive intelligent traffic control algorithm using multi-agent RL based on improved Friend Deep Neural Network Q-learning, namely, Friend-DQN~\cite{7}. The Friend-DQN method does not increase joint state-action space exponentially as the number of intersections increases, and thus,  Friend-DQN will converge faster than traditional Q-learning and DQN.
More specifically, the main contributions of this paper are:
\begin{itemize}
\item	We propose a Friend-DQN model to deal with multi-intersection problems, which aims to minimize average vehicle waiting time;
\item	We compare the Friend-DQN model with fixed-phase, centralized DQN and independent DQN for different numbers of intersections in terms of convergence speed and average waiting time;
\item	We justify the effectiveness and superiority of the Friend-DQN model on the SUMO platform.
\end{itemize}

\section{Related Work}

The first academic application of RL technique in a traffic signal control problem was the successful application of SARSA to traffic signal control~\cite{4,11,12}. SARSA is an on-policy algorithm for learning a Markov decision process policy, which combines timely and intelligent traffic control policies with real-time road traffic~\cite{26}. Srinivasan et al.~\cite{13} uses a distributed multi-agent model to solve the traffic signal control problem, where each agent has an independent Q table to learn and judge the execution phase. Experiments demonstrate the effectiveness of Q learning.

Recently, IntelliLight was proposed to be implemented using DQN and tested in a real road network~\cite{14}. IntelliLight is combined with a specific traffic signal control problem. The environment consists of traffic signal phases and traffic conditions, and the state is a characteristic representation of the environment information. The agent inputs the state and controls the signals as an action, such as changing the traffic signal phase or the duration of the signal, and then the agent gets a reward from the environment. The agent in IntelliLight implements this through a DQN network, which updates the model based on the loss function of the DQN network to maximize the reward. It is worth noting that the research argues that the agent has to analyze and understand the strategy in the context of the actual scenario. There is no denying that IntelliLight does perform well in real cities. However, it is still a centralized RL algorithm, which means that it still cannot avoid the vast space and time occupation when dealing with a large-scale network.

\begin{figure}[t]
  \centering
  \includegraphics[width=1\linewidth]{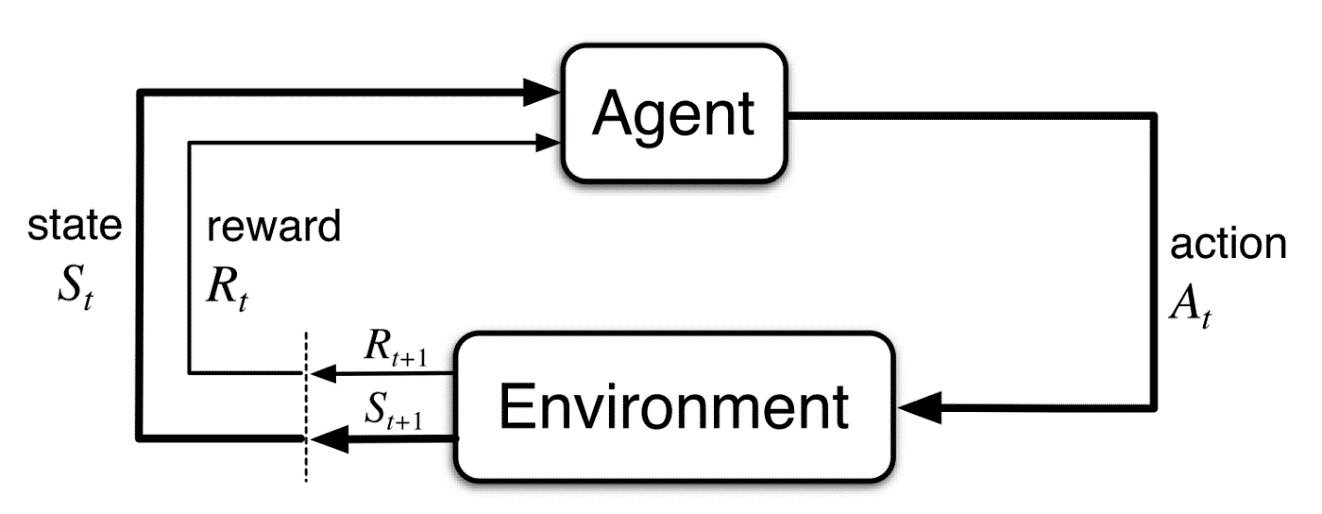}
  \caption{Interaction of agents with their environment in Markovian decision processes.}
  \label{fig1}
\end{figure}

A multi-agent deep RL method that combines the DQN algorithm with transfer planning can solve the difficulty of centralized RL~\cite{15}. Transfer planning can avoid the problems of previous multi-agent reinforcement learning i.e. space and time occupation and allow for faster and more scalable learning. This study introduces a new reward function to the ATSC problem. It solves the problem of extreme delays previously caused by a single average vehicle waiting time as a reward by combining criteria such as transport penalties and vehicle delays with different weights to calculate a new reward. Finally, the control of multi-agent is achieved by transfer planning and max-plus coordination algorithms. This approach reduces the problem of large spaces for single agents to some extent, but it is still precarious and sometimes underperforms due to the use of deeper networks.

Recent research has proposed the use of independent advantage actor-critic (A2C) for traffic signal control instead of Q-learning~\cite{5}. Although they expanded the state representation by including observations and fingerprints of neighbouring agents in each agent's state and used a spatial discount factor to adjust the global reward for each agent, they did not consider the higher-order relationships of the agents. Others have used the more robust DDPG instead of the A2C method. However, in the past DDPG-based traffic control frameworks ~\cite{24,25} focused only on single intersections and could not be applied to large-scale traffic networks.

\begin{table}[t]\centering
\normalsize
  \caption{THE PARAMETERS OF FRIEND-DQN}
  \label{tab:locations}
  \begin{tabular}{ll}\toprule
    \textit{Parameter} & \textit{Value} \\ \midrule
    Greedy rate & 0.9 \\
    Maximize epsilon value & 0.9 \\
    Learning rate & 0.1 \\
    Memory size & 20000 \\
    Batch size &50 \\ 
    Target network weight updating frequency &400 \\ \bottomrule
  \end{tabular}
\end{table}
\section{Methodology}
To implement more robust adaptive traffic signal control, we propose the decentralized Friend-DQN algorithm in the framework of reinforcement learning. Since the theory of Friend-learning is an enhancement of Nash-learning, we firstly briefly review fundamentals of MDP and Nash-Q before introducing Friend-learning.

\begin{figure*}[ht]
\centering
  \includegraphics[width=1\linewidth]{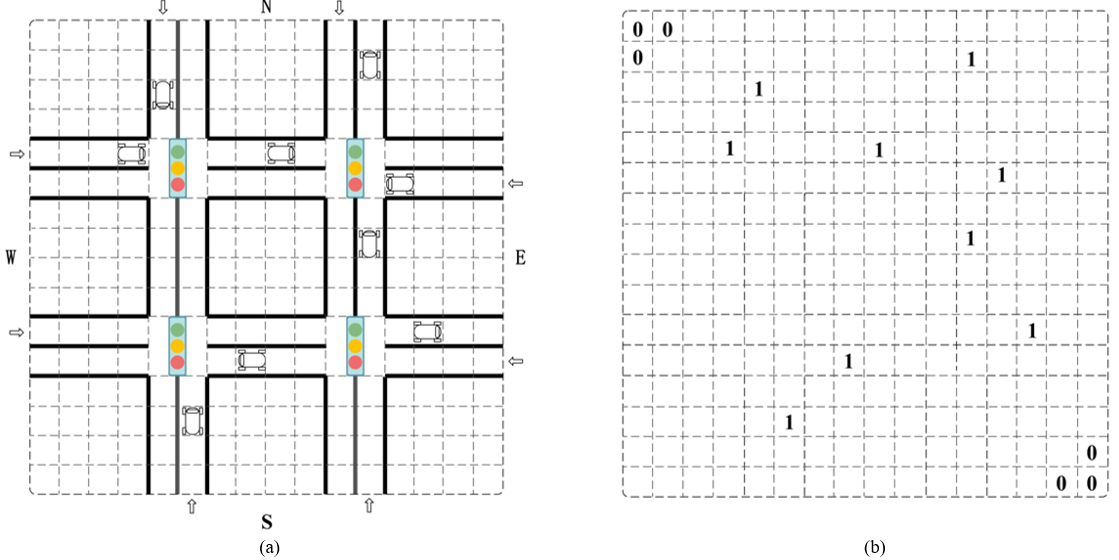}
  \caption{a) Traffic situation. b) Simplified example of state representation in a 16×16 matrix.}
  \label{fig2}
\end{figure*}

\subsection{Model}

It can be observed from some literature~\cite{15,16,17,18} that traffic signal control problem can be described as a Markov Decision Process (MDP). MDP aim to detect the state of the environment, select actions, and associate goals related to the state of the environment in a simple form. The definition of the MDP contains the state space $S$, action set $A$, transition probability $P$ and reward $R$. The agent reacts to an environmental state $s_t\in S$ by taking a possible action $a_t\in A$. It ends up in state $s_{t+1}$ with some transition probability $p(s_{t+1}|s_t,a_t)\in P$ and receives a reward signal $r(s_t,a_t,s_{t+1})\in R$~\cite{19}. The process is shown in Figure \ref {fig1}.

\noindent\textbf{State S}: Figure \ref {fig2}a shows a four-intersection traffic signal control problem, where the matrix represents an image in SUMO to represent the state of the vehicles around the intersections. In figure \ref {fig2}b, referring to the definition as Tobias~\cite{20}, we use a matrix to represent the vehicle position information in the traffic signal controlled lane as the state. The whole two-dimensional space shown in figure \ref {fig2}a is split into several squares with the same length and width, each of which is filled with one or zero representing the existence of vehicles.

\noindent\textbf{Action A}: At each step, agents can choose a traffic signal duration as the action which will change the states. In our design, there are six actions per junction, which are six different phases in five-second intervals from 10 to 30 seconds and can be selected as the phase of the junction traffic signal at each update.

\noindent\textbf{Transition probability P}:  Transition probability $p(s_{t+1}|s_t,a_t)$ defines the probability of state transition from the current $s_t$ to the next state $s_{t+1}$ when the agent takes action $a_t$.

\noindent\textbf{Reward R}:  Reward is defined as the difference the average waiting time of vehicles between the next state and the current state.

The rewards that define the traffic signal control problem are not uniform. Here we take the metric of the reduced waiting time for vehicles at intersections. However, in a real traffic road network, the average waiting time will be calculated when a vehicle finishes its journey. This leads to severe latency problems. So here we select the action at time t and calculate the reward and learning at the next phase, i.e., time $t+1$.

\begin{figure}[t]
  \centering
  \includegraphics[width=1\linewidth]{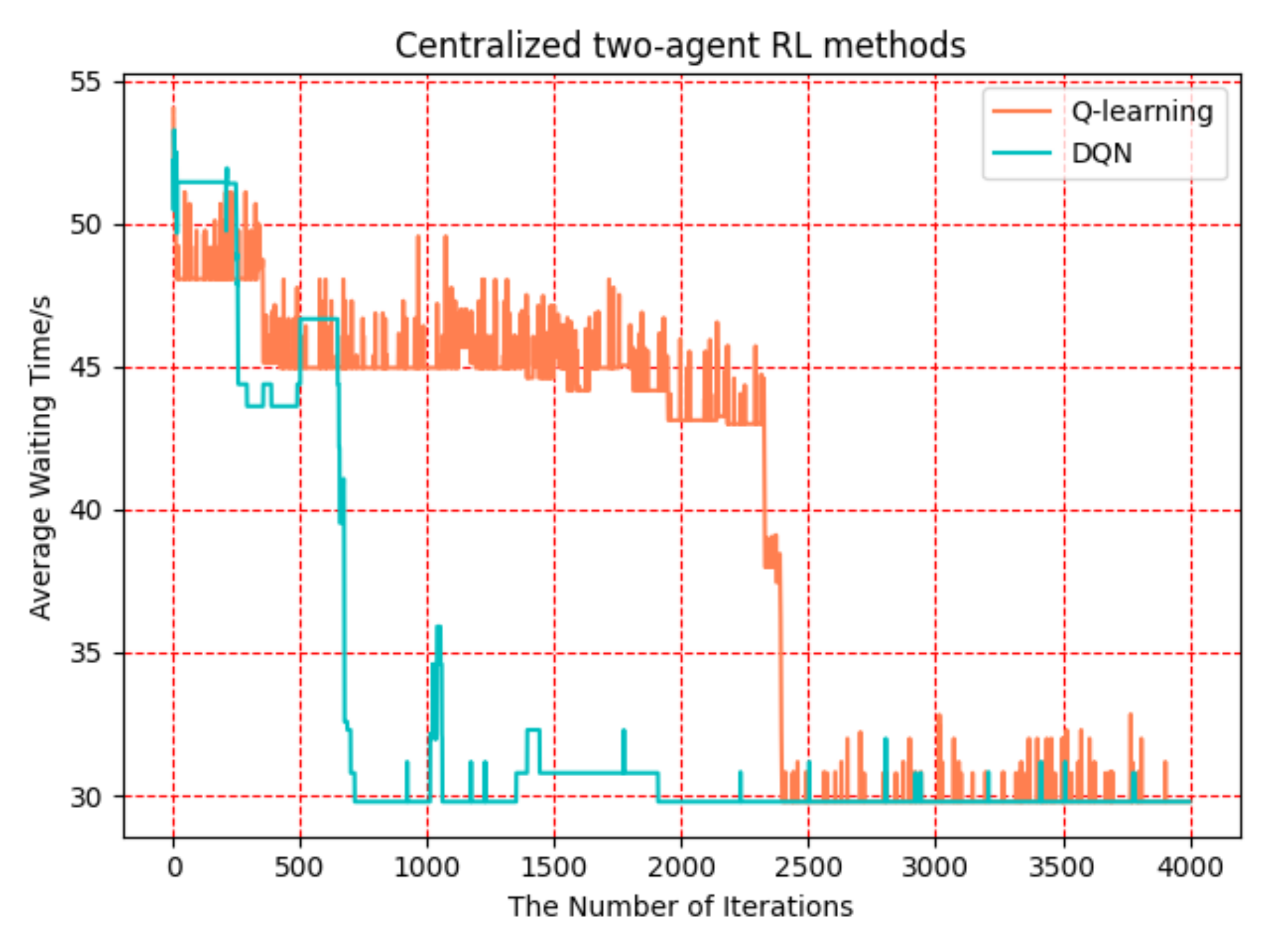}
  \caption{Centralized two-agent RL method.}
  \label{fig3}
\end{figure}
The final reward rt for each time step is as follows:
\begin{eqnarray}
{{r}_{t}} & = &
\underset{n=1}{\overset{N}{\mathop \sum }}\,{{w}_{t}}-\underset{n=1}{\overset{N}{\mathop \sum }}\,{{w}_{t+1}}
\end{eqnarray}
where $N$ represents the number of vehicles on the lanes, and $w$ is the waiting time.

The objective of the agent is to maximize the accumulation of rewards. By formalizing the reward, it is passed from the environment to the agent. The agent is rewarded with the sum of the rewards:
\begin{eqnarray}\label{eq:vcg}
{{G}_{t}} & = &
{{r}_{t+1}}+{{r}_{t+2}}+{{r}_{t+3}}+...+{{r}_{T}}
\end{eqnarray}

To make the agent more "farsighted", i.e., to consider future rewards, introduce a discount factor  $gamma$, then the agent chooses action $A_t$ at time $t$ to maximize the desired discounted reward: 
\begin{eqnarray}\label{eq:vcg}
{{G}_{t}} & = &
{{r}_{t+1}}+\gamma {{r}_{t+1}}+{{\gamma }^{2}}{{r}_{t+2}}+...=\underset{0}{\overset{\infty }{\mathop \sum }}\,{{\gamma }^{k}}{{r}_{t+k+1}}
\end{eqnarray}

If $gamma$ is equal to 0, then the agent only considers current rewards and the objective of the agent is to learn how to choose an action $A_t$ to maximize $r_t$.

\subsection{Nash-Q and Friend-Q algorithm}
The Multi-agent Nash Q algorithm considers other agents when selecting actions, i.e., it selects the action that makes the system reward and largest. 
\begin{figure}[t]
  \centering
  \includegraphics[width=1\linewidth]{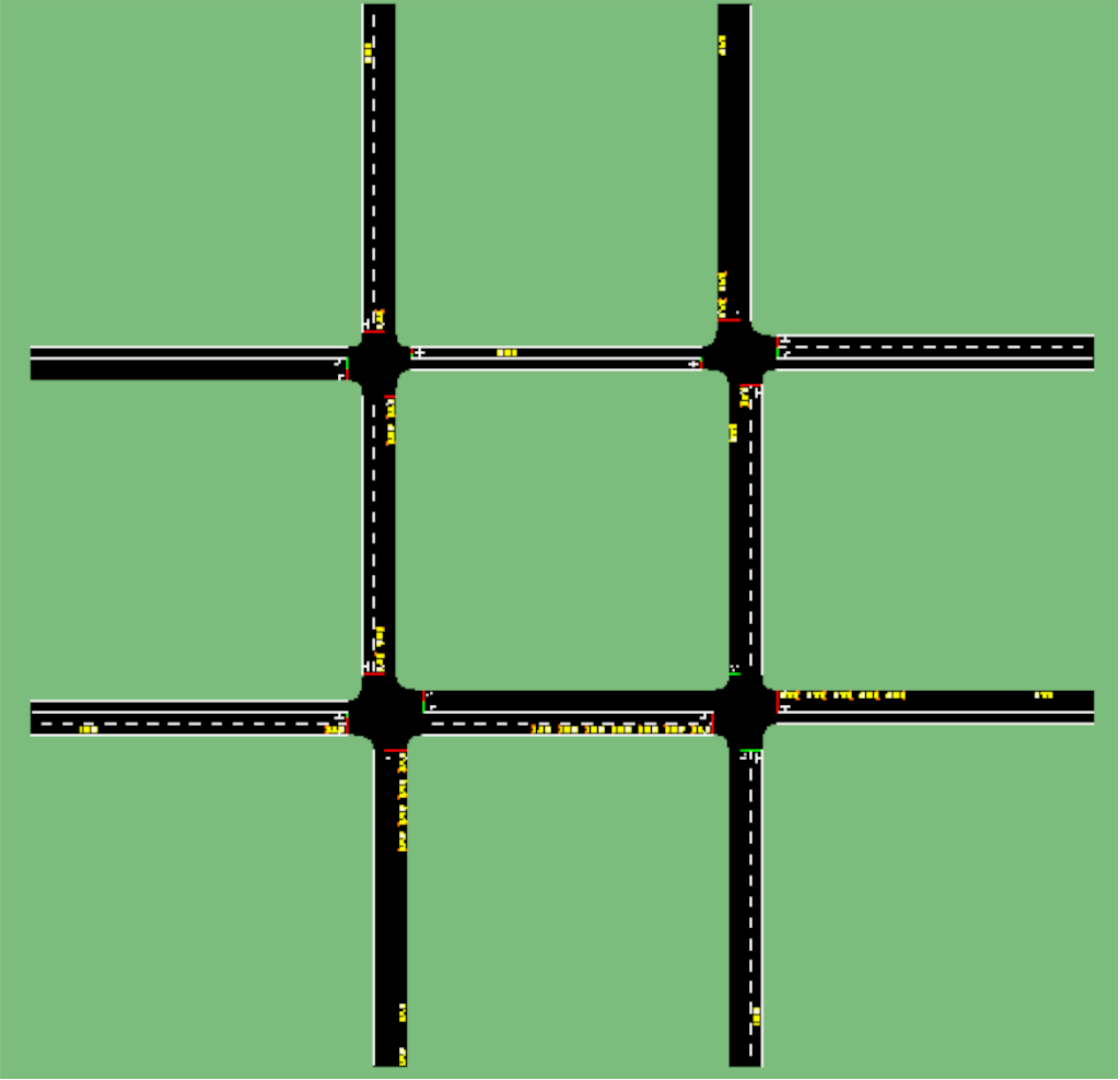}
  \caption{Traffic network with 4 intersections.}
  \label{fig4}
\end{figure}
The Nash equilibrium locks in each agent's strategy because it cannot simply change its own strategy to increase its payoffs~\cite{22}. The learning agent, indexed by $i$, learns its Q-value by making arbitrary guesses at moment $t$. At moment $t$, agent $i$ takes an action by observing the current state. Afterward, it learns the reward of itself, the actions taken by all other agents, the rewards of others and the new state $s'$. A Nash equilibrium is then calculated for the current phase and updated the Q-value according to: 
\begin{eqnarray}\label{eq:vcg}  
\begin{split}
Q^i_{t+1}(s,a^1,...,a^n) & = & (1-\alpha)Q^i_t(s,a^1,...,a^n)\\ 
&  + & \alpha_t[r^1_t+\beta NashQ^i_t(s')]
\end{split}
\end{eqnarray}
Where  %
\begin{eqnarray}\label{eq:vcg}
NashQ_{t}^{i}\left( {{s}'} \right) & = & {{\pi }^{1}}\left( {{s}'} \right)...{{\pi }^{n}}\left( {{s}'} \right)Q_{t}^{i}\left( {{s}'} \right)
\end{eqnarray}

Its updates are asynchronous, that is, only actions relating to the current state are updated.

The convergence condition for the Nash Q-Learning algorithm to converge in a cooperative or adversarial equilibrium setting is that a global optimum or saddle point can be found in each state s of the stage game. The Nash Q-learning algorithm can only converge if this condition is satisfied.

Nevertheless, in a transport network, the relationship between different agents is not just competitive, we want multiple agents to work together to get the most rewards for the whole transport system. We, therefore, refer to the Friend or Foe Q-Learning (FFQ) proposed by Littman~\cite{10}. Friend-Q assumes that the opponent is like a friend who maximizes everyone's benefits, so add the action space of player B to Q.
\begin{eqnarray}\label{eq:vcg}
FriendQ_{t}^{i}\left( s,{{Q}_{1}},{{Q}_{2}} \right)=\underset{{{a}_{1}}\in {{A}_{1}},{{a}_{2}}\in {{A}_{2}}}{\mathop{max}}\,Q\left[ s,a1,a2 \right]
\end{eqnarray}

By appropriately replacing NashQ with FriendQ, distributed agent systems can achieve a balanced strategy through cooperative learning. 

\begin{table}[t]\centering
\normalsize
  \caption{THE ATTRIBUTES ON SUMO}
  \label{tab:locations}  \begin{tabular}{ll}\toprule
    \textit{Attributes} & \textit{Value} \\ \midrule
    Number of junctions & 2/3/4 \\
    Number of roads & 12 \\
    Average length of lanes & 100m \\
    Number of lanes per road & 3 \\
    Phase duration & 40s \\ 
    Arrival distribution &	Uniform \\
   Simulation duration & 5.5hours \\ \bottomrule
  \end{tabular}
\end{table}

\begin{algorithm}[!h]
    \caption{Multi-Agent Friend-DQN}
    \label{alg:FDQN}
    \renewcommand{\algorithmicrequire}{\textbf{Input:}}
    \renewcommand{\algorithmicensure}{\textbf{Output:}}
    \begin{algorithmic}[1]
        \STATE  initialization
        \STATE  Let $t = 0$, get the initial state $s_0$.
        \STATE  Let the learning agent be indexed by $k$.
        \FOR{all $s \in S$ and $a_i \in A_i$, $i = 1,...,n$ }
          \STATE Let $Q^i_t(s, a^1,...,a^n) = 0$.
        \ENDFOR
        \LOOP
          \STATE Choose action $a^k_t$.
          \STATE Observe $r^1_t,..., r^n_t; a^1_t,..., a^n_t$,and $s_{t+1} = s'$.
          \FOR{i = 1,...,n}
           \STATE $Q^i_t(s,a^1,...,a^n) = (1-\alpha)Q^1_t(s,a^1,...,a^n) +\alpha_t[r^i_t + \beta  FriendQ^i_t(s,Q_1,...,Q_n)]$
          \ENDFOR
          \STATE Where $\alpha \in (0, 1)$ is the learning rate, and $FriendQ^i_t(s,Q_1,...,Q_n)$ is defined in (8)
          \STATE Let $t := t + 1$
        \ENDLOOP
    \end{algorithmic}
\end{algorithm}
\subsection{Multi-agent Friend-DQN algorithm}
\begin{figure*}[ht]
\centering
  \includegraphics[width=1\linewidth]{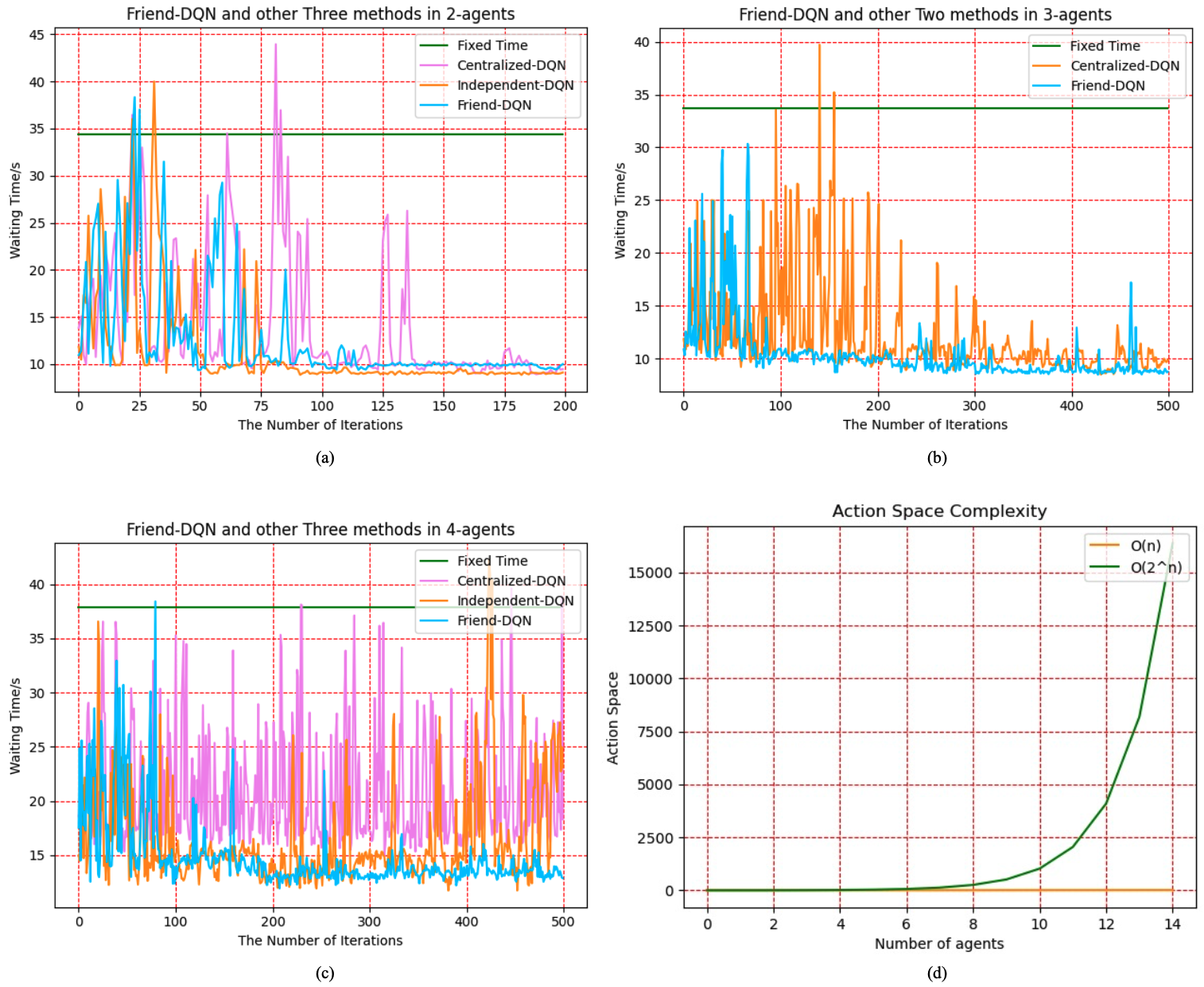}
  \caption{a) 4 methods in 2-agents. b) 3 methods in 3-agents. c) 4 methods in 4-agents. d) Action Space Complexity for DQN and Friend-DQN}
  \label{fig5}
\end{figure*}
However, this is not enough, we also need to extend Friend-DQN so that it can be adapted to larger systems of agents. The Friend-Q value becomes as follows:
\begin{small}
\begin{eqnarray}
\label{eq:vcg}
\begin{split}
  & Frien{{d}_{i}}\left( s,{{Q}_{1}},\ldots ,{{Q}_{n}} \right) = \\ 
 & \underset{\pi \in \text{ }\!\!\Pi\!\!\text{ }\left( {{X}_{1}} \cdot \cdot \cdot {{X}_{k}} \right)}{\mathop{max}}\,\underset{{{x}_{1}},...,{{x}_{k}}\in {{X}_{1}}\cdot\cdot\cdot{{X}_{k}}}{\mathop \sum }\,\pi \left( {{x}_{1}} \right)...\pi \left( {{x}_{k}} \right)Q\left[ s,{{x}_{1}},...,{{x}_{k}} \right] \\ 
 \end{split}
\end{eqnarray}
\end{small}
FriendQ is a strategy for improving Q-learning to find equilibrium for multi-agent systems. Q-learning needs to generate Q-tables in runtime, so when processing traffic, the large state space can result in the need to generate a colossal q-table. We examined the ATSC of DQN and Q-learning at two junctions and figure \ref {fig3} showed that DQN converges much faster than Q-learning due to the neural network it introduces. Therefore, it is necessary to improve Friend-DQN to increase the convergence speed further.

DQN uses neural networks to represent Q values, which is what becomes represented by Q networks. We use the target Q value as a label to get the Q value to converge to the target Q value ~\cite{21}. 

Therefore, the loss function for Q-network training is:
\begin{eqnarray}\label{eq:vcg}
L\left( w \right) & = & E[{{(r+\gamma \underset{{{a}'}}{\mathop{max}}\,Q\left( {s}',{a}',w \right)-Q\left( s,a,w \right))}^{2}}]
\end{eqnarray}

Where $r+\gamma \underset{{{a}'}}{\mathop{max}}\,Q\left( {s}',{a}',w \right)$   is the target.

We determined the loss function, i.e., cost, and the way to obtain the samples, the whole algorithm of DQN is shaped.

where X represents the set of all agents.

The hyperparameters of Friend-DQN are shown in TableI. Algorithm\ref{alg:FDQN} illustrates our proposed algorithm.

\section{EXERIMENTAL RESULTS}
Here we choose the SUMO platform for the simulation. Using four traffic lights as an example, see Figure \ref {fig4}, all eight traffic roads will depart at a specific frequency. The system configuration parameters are shown in TableII.

We compared Friend-DQN with a traditional centralized DQN, independent-DQN and a fixed-time ATSC on SUMO. Because the traffic network is asymmetric when there are three agents, the independent method cannot be directly experimented with the traffic network. Thus we compared Friend-DQN with independent-DQN in the two-agents and four-agents settings. The results are shown in Figure \ref {fig5}, which shows that the fixed time is not optimized for traffic. Although the traditional DQN can optimize ATSC, the convergence speed is much slower than the decentralized Friend-DQN algorithm. Moreover, as more junctions are added, the gap between Friend-DQN and centralized DQN grows. At two agents, the convergence speed of independent-DQN and Friend-DQN is approximated. This is due to that the action spaces of both methods are the same. However, Figure \ref {fig5}(c) shows that the indenpendent-DQN method keeps oscillating unable to converge to a policy when there are four agents.This is because there is no communication between agents and cannot converge to a stable policy.

In addition, Figure \ref {fig5}(c) shows that after 500 trailing epochs at four traffic junctions, the DQN still does not converge. When we analyze the complexity of the action space of these two algorithms, we can see that Friend-DQN outperforms centralized RL to a large extent. Action space represents the projection of the actions in the system. The action space complexity of centralized RL is $O(2^n)$, while the action space complexity of Friend-DQN is only $O(n)$. So when there are four junctions, the centralized approach has 1296 action choices compared to 24 for Friend-DQN. This is why there is such a big difference between the two methods in convergence speed.

Figure \ref {fig5}(d) shows that as the number of traffic junctions increases, the time and space required for centralized algorithms will become a huge hassle. That is why we are developing decentralized multi-agent to implement ATSC.

\section{Conclusions}
In this paper, we have demonstrated that Friend-DQN is a promising approach to adaptive traffic signal control. In an entire traffic network, traffic flows are dynamic and change over time. Our proposed approach is based on information in a phase statistic and learning the optimal joint action of multiple agents in different situations. Vehicle location information and queue length are used as collected information. Many states and joint actions are learned in training, so our algorithm can be extended to more extensive traffic networks.

At the same time, the decentralized Friend-DQN algorithm is a scalable multi-agent approach to deep reinforcement learning. Using cooperation to achieve equilibrium avoids the problems of single-agent reinforcement learning and allows for faster and more scalable learning. Its action space complexity is linear, so as more traffic intersections are added, the algorithm performs significantly better than earlier single-agent traffic signal control efforts. 

We conducted simulation experiments on SUMO for four junctions and proved the performance and stability of the algorithm. Simulation results also illustrate that our approach outperforms the fixed-time and DQN algorithms in different traffic conditions.

Our current system only considers cooperation between traffic junctions. It can lead to some inequities i. e. excessive waiting times at certain junctions. Future work includes applying the Nash Equilibrium and Friend-DQN so that agents can cooperate and also secure their interests.

\section*{Acknowledgment}
This work was supported in part by XJTLU Research Development Funding RDF-21-02-015. Any comments should be address to Dr. Wang Shangbo.
 
\bibliographystyle{ieeetr}  
\bibliography{sample}

\end{document}